 \documentclass[preprint,3p,times,twocolumn]{elsarticle}
\usepackage{amssymb}
\usepackage{amsmath}
\usepackage{algpseudocode}
\usepackage{algorithm}
\usepackage[dvipsnames]{xcolor}
\usepackage{longtable}
\usepackage{url}
\usepackage{booktabs}
\usepackage{adjustbox}
\usepackage{comment}

\usepackage{titlesec}

\titlespacing*{\section}
  {0pt}{1ex plus 1ex minus .2ex}{0.8ex plus .1ex}

\titlespacing*{\subsection}
  {0pt}{0.8ex plus 1ex minus .2ex}{0.6ex plus .1ex}

\journal{Pattern Recognition Letters}

\begin{document}

\begin{frontmatter}

%% Title, authors and addresses

%% use the tnoteref command within \title for footnotes;
%% use the tnotetext command for theassociated footnote;
%% use the fnref command within \author or \affiliation for footnotes;
%% use the fntext command for theassociated footnote;
%% use the corref command within \author for corresponding author footnotes;
%% use the cortext command for theassociated footnote;
%% use the ead command for the email address,
%% and the form \ead[url] for the home page:
%% \title{Title\tnoteref{label1}}
%% \tnotetext[label1]{}
%% \author{Name\corref{cor1}\fnref{label2}}
%% \ead{email address}
%% \ead[url]{home page}
%% \fntext[label2]{}
%% \cortext[cor1]{}
%% \affiliation{organization={},
%%             addressline={},
%%             city={},
%%             postcode={},
%%             state={},
%%             country={}}
%% \fntext[label3]{}

\title{Topological Data Analysis for Unsupervised Anomaly Detection and Customer Segmentation on Banking Data}

%% use optional labels to link authors explicitly to addresses:
%% \author[label1,label2]{}
%% \affiliation[label1]{organization={},
%%             addressline={},
%%             city={},
%%             postcode={},
%%             state={},
%%             country={}}
%%
%% \affiliation[label2]{organization={},
%%             addressline={},
%%             city={},
%%             postcode={},
%%             state={},
%%             country={}}

\author[a,b]{Leonardo Aldo Alejandro Barberi}\ead{lbarberi@ethz.ch} %% Author name
\author[b]{Linda Maria De Cave}\ead{lindamaria.decave@imtf.com}

%% Author affiliation
\affiliation[a]{organization={Department of Computer Science, ETHZ},%Department and Organization
            addressline={Universitätsstrasse 6}, 
            city={Zürich},
            postcode={8092}, 
            state={Zürich},
            country={Switzerland},
            }
\affiliation[b]{organization={AI Research, IMTF SA},%Department and Organization
            addressline={Förrlibuckstrasse 190}, 
            city={Zürich},
            postcode={8005}, 
            state={Zürich},
            country={Switzerland},
            }

%% Abstract
\begin{abstract}
%% Text of abstract
This paper introduces advanced techniques of Topological Data Analysis (TDA) for unsupervised anomaly detection and customer segmentation in banking data. Using the Mapper algorithm and persistent homology, we develop unsupervised procedures that uncover meaningful patterns in customers' banking data by exploiting topological information. The framework we present in this paper yields actionable insights that combine the abstract mathematical subject of topology with real-life use cases that are useful in industry.
\end{abstract}

%%Graphical abstract
%\begin{graphicalabstract}
%\includegraphics[width=2.2\linewidth]{graphical_abstract.pdf}
%\end{graphicalabstract}

%%Research highlights
%\begin{highlights}
%\item Topological data analysis methods offer expressive solutions to perform exploratory data analysis on high-dimensional banking data. 
%\item Graph-based anomaly detection techniques leveraging the Mapper algorithm are capable of identifying isolated or weakly connected nodes as potential fraud cases without relying on labeled data.
%\item The Mapper graph's topological information is valuable beyond anomaly detection. We present an outline for a community detection algorithm that exploits this property and yields statistically significant customer segments.
%\end{highlights}

%% Keywords
\begin{keyword}
%% keywords here, in the form: keyword \sep keyword
Graph Pattern Recognition, Anomaly Detection, Topological Data Analysis, Mapper Algorithm
%, Persistent Homology
%% PACS codes here, in the form: \PACS code \sep code

%% MSC codes here, in the form: \MSC code \sep code
%% or \MSC[2008] code \sep code (2000 is the default)

\end{keyword}

\end{frontmatter}

%% Add \usepackage{lineno} before \begin{document} and uncomment 
%% following line to enable line numbers
%% \linenumbers

%% main text
%%

%% Use \section commands to start a section
\section{Introduction}
\label{sec:intro}

The detection of financial anomalies has been a persistent challenge for financial institutions, with documented cases dating back over two millennia. The methods employed by fraudsters are continuously evolving and, as they become increasingly sophisticated, financial institutions have to adapt their anomaly detection strategies.

Some of the difficulties faced by financial institutions in anomaly detection are due to the vast volume of transactions, the high dimensionality of transactional data, and the scarcity of labeled fraud cases~\cite{bahnsen2016, dalpozzo2015, jurgovsky2018}. Traditional machine learning approaches often struggle with these issues, as they typically require labeled data for training and may not generalize well to new, unseen fraud patterns~\cite{{ngai2011}}. Furthermore, these models often rely on overly simplistic assumptions about the distribution of the data, which may fail to consider the complex nature of customer behaviour~\cite{bhattacharyya2011}.

Unsupervised customer segmentation remains a key challenge for financial institutions too. Classical clustering algorithms have been applied to this task~\cite{mozumder2024optimizing, John2023}, but they rely on assumptions about cluster shape, size, and number that cannot be made in certain use cases.
%Community detection offers a natural alternative that overcomes these limitations~\cite{Nasirzonouzi2025}, though careful attention must be paid to scalability issues.

Topological Data Analysis (TDA) provides a framework for understanding high-dimensional data through its shape and connectivity, offering an alternative for unsupervised learning tasks. The two foundational tools in TDA used in this paper are the Mapper algorithm and Persistent Homology~\cite{singh2007topological, edelsbrunner2010computational}. 
%Readers interested in a deeper understanding of these core techniques are referred to~\cite{singh2007topological} and~\cite{edelsbrunner2010computational}.

The Mapper algorithm is used to transform transactional data into a network. Unlike traditional clustering or classification methods, this algorithm provides insights into the global structure of the data, helping to uncover relationships and patterns that may not be evident in standard feature-space analyses. The topological information encoded in the Mapper graph can be leveraged using persistent homology, a mathematical framework that provides insights by tracking the evolution of structures across multiple scales~\cite{edelsbrunner2010computational}. 

%The Mapper algorithm works by first mapping the data through a chosen filter function, which defines a lower-dimensional representation of the dataset. The mapped data is then covered with overlapping hypercubes, and, within the preimage of each hypercube, the data is clustered using a clustering algorithm. Clusters that share data points across intervals are connected to form a graph, referred to as the \emph{Mapper graph}~\cite{singh2007topological}.
%This representation allows for the visualization and exploration of complex data structures while preserving important topological features~\cite{singh2007toplogicalmapper}. 

%The structure of the Mapper graph depends on several parameters that must be carefully selected, of which the filter function, the clustering algorithm, its hyperparameters, and further parameters that define the open interval cover. Beyond parameter selection, the Mapper algorithm must also be assessed in terms of its ability to reveal meaningful patterns in transactional data as, in the context of anomaly detection and customer segmentation, an appropriate algorithm should be capable of identifying anomalous behaviour and meaningful segments, even in the absence of labels. 

Our objective is to address the following research questions using a TDA-based framework. 
\\\textbf{Research Question 1.} In the absence of labeled data, how can the Mapper parameter selection process be optimized to ensure the robustness of our approach? 
\\\textbf{Research Question 2.} Can the Mapper algorithm effectively reveal patterns in transactional data that indicate anomalous behaviour? 
\\\textbf{Research Question 3.} How can the Mapper graph's topological information be used to perform customer segmentation?

The paper is structured as follows. We begin by describing our approach, focusing on key aspects such as the optimization of Mapper parameters, the design of an anomaly detection method based on graph connectivity, and we briefly outline a persistence-driven community detection algorithm that exploits the Mapper graph's intrinsic information to perform customer segmentation. The results chapter presents the experimental findings, with a focus on evaluating the effectiveness of these methods in detecting anomalies and segmenting customers. Finally, we present potential directions for future research.

\section{Related Work}
\label{sec:related_work}

Anomaly detection using TDA has been previously developed in~\cite{binshiraj2024topologicalanomalydetection}, where isolated Mapper graph nodes are classified as outliers. We extend this approach by leveraging intrinsic Mapper information rather than treating it exclusively as a graph, refining the definition of isolated nodes. Additionally, we enhance robustness by aggregating classifications across a set of stable Mapper graphs. While classification ensembles are a tried and tested technique~\cite{Rokach2010ensemblelearning}, to our knowledge, we are the first to apply it in the context of classification using the Mapper graph.

One major limitation of the Mapper algorithm is its sensitivity to design choices and parameters. To address this, various methods have been developed to guide parameter selection across different use cases. Ensemble learning techniques, as presented in~\cite{fitzpatrick2023ensemblelearningmapper} and~\cite{kanglim2021ensemblemapper}, attempt to mitigate these effects by aggregating multiple Mapper graphs, computed with different parameters, into a single graph output. While promising, these approaches are computationally expensive, and their scalability to large, complex datasets remains uncertain. 

Probabilistic Mapper architectures, such as the differentiable Mapper, have been proposed to automate parameter selection~\cite{oulhaj2024differentiablemappertopologicaloptimization}. While promising in theory, they lack experimental validation for use cases such as the one we present in this paper. Similarly, statistical guarantees for certain parameter choices have been established~\cite{carrière2017statisticalanalysisparameterselection}, but these rely on assumptions about the underlying topological space that we cannot make. 

The challenge of Mapper parameter selection was tackled through a stability optimization routine (see Section~\ref{sec:stability_optimization}), as presented in~\cite{byrne-etal-2022-topic}. 

While not central to our paper, we also present an outline for a community detection algorithm using the Mapper graph. Although community detection is a well-studied problem, we have developed a novel persistent-homology-driven approach. Persistence-based graph clustering methods have demonstrated expressiveness comparable to state-of-the-art models~\cite{ballester2024expressivitypersistenthomologygraph}. 
In Chapter~\ref{sec:results_cust_seg}, we compare our method against ToMATo (Topological Mode Analysis Tool), a persistence-driven clustering algorithm~\cite{chazal2011}. 

\section{Methodology}
\label{sec:methodology}

In this section, we present procedures to guide the design choices for constructing the Mapper graph and performing anomaly detection and customer segmentation. 

%Figure~\ref{fig:pipeline} presents an overview of the methodology.

%\begin{figure}
%    \centering
%    \includegraphics[width=\linewidth]{pipeline_overview.png}
%    \caption{Overview of the pipeline for anomaly detection and customer segmentation.}
%    \label{fig:pipeline}
%\end{figure}

\subsection{Mapper Parameter Selection}
\label{sec:stability_optimization}
The Mapper algorithm maps high-dimensional data to a lower dimensional space using a \emph{filter function}. An open cover of the image space is constructed and, within the pre-image of each interval of the open cover, the data is clustered using a clustering algorithm. Each cluster is represented as a node in the output \emph{Mapper graph}, and clusters that share elements with clusters from other interval pre-images are connected by an edge~\cite{singh2007topological}. 

This algorithm requires a careful selection of design parameters, including the filter function, a clustering algorithm and cover hyperparameters. For our use case, the filter function was chosen to be the projection onto the first two principal components of the dataset, a common practice in Mapper literature. 

Due to its topological nature, AuToMATo presents itself as an ideal clustering algorithm~\cite{huber2024automatooutoftheboxpersistencebasedclustering}. Built on ToMATo, a persistence-driven clustering algorithm, AuToMATo automates the choice of a central design parameter using a statistical technique called \emph{bottleneck bootstrapping}. With respect to ToMATo, this choice enables us to restrict the search for optimal clustering-related parameters to only one parameter $k$ that determines $k$-nearest neighbours graph used to estimate topological connectivity in AuToMATo, and is influential in the connectivity of the output Mapper graph. Readers wishing to learn more about the role of this parameter are referred to~\cite[Section 3]{chazal2011}.

Following~\cite{byrne-etal-2022-topic}, the Mapper hyperparameters were selected to optimize stability. 
%Without ground truth labels, stability optimization ensures that unsupervised learning results remain robust under data perturbations.
% In our case, aside from being useful due to the absence of ground truth labels, stability optimization guarantees that the mapper graph calculated on a subsample of the entire dataset can be considered to be a faithful representation of the entire topological space.
In order to define a cover of the image space of the filter function, two parameters are needed, the \emph{gain} $g$ that controls the overlap between two consecutive intervals of the open cover, and the \emph{resolution} $r$ that controls the measure of each open interval.

Let $\mathcal{M}_\theta(D)$ represent the Mapper graph on a dataset $D$ with parameters $\theta$. Equipped with a distance measure $d(\mathcal{M}, \mathcal{M}')$, the \emph{instability} of $\mathcal{M}$ with parameter $\theta$ can be quantified as the expected distance between $\mathcal{M}_\theta(D)$ and $\mathcal{M}_\theta(D')$, where $D$ and $D'$ are data samples obtained from the same data generating process. Concretely, the \emph{instability score} is defined as 
$$
    S(\mathcal{M}_\theta, d) = \frac{2}{n(n-1)}\sum_{i=0}^n\sum_{j=i+1}^n d(\mathcal{M}_\theta(D_i), M_\theta(D_j))
$$

where $D_i$ and $D_j$, for $1\leq i,j\leq n$, are independent samples from the dataset $D$~\cite{byrne-etal-2022-topic}.

In our use case, the parameters $\theta \in \Theta$, where $\Theta=\{(g, r, k): g\in [0, 1], r \in \mathbb{R}_+, k \in \mathbb{N}\}$ and the choice of graph distances was restricted to the \emph{spectral distance} and the \emph{NetSimile distance}~\cite{byrne-etal-2022-topic, berlingerio2012netsimilescalableapproachsizeindependent}. These two graph distances were chosen for their complementary scope: the spectral distance captures overarching graph structures, while the NetSimile distance focuses on local-level network features.

The stability optimization routine yields a stable region $\Theta_S$ for the set of parameters, where 
$$\Theta_S = \{\theta \in \Theta: S(\mathcal{M}_\theta, d_N) <\epsilon_N, S(\mathcal{M}_\theta, d_s) <\epsilon_s\}$$
for two $\epsilon_N, \epsilon_s > 0$ of our choice. This confidence region plays a key role in guiding our parameter selection process, though it does not directly determine a single best-performing parameter choice. As will be discussed in Section~\ref{sec:ensemble_anomaly_detection}, this is not a concern for anomaly detection, as we consider multiple Mapper graphs in the stable region. 
%However, for customer segmentation, the optimal parameter within the stable region $\theta \in \Theta_S$ was chosen based on visual analysis.

\subsection{Anomaly Detection}
\label{sec:anomaly_detection_meth}

In this section, we present the algorithm used to detect customers with anomalous transactional behaviour. In the Mapper graph, each node represents a cluster of the pre-image of an open interval in the cover of the image space, and two nodes are connected if their corresponding clusters share points. The goal of our anomaly detection algorithm is to identify nodes corresponding to clusters of customers with unusual behaviour.

To this end, the aim is to identify isolated nodes of the Mapper graph. Rather than relying solely on node degree to determine isolation, as is done in~\cite{binshiraj2024topologicalanomalydetection}, we also incorporate node-level features extracted by NetSimile~\cite[Section 2]{berlingerio2012netsimilescalableapproachsizeindependent}. These features provide a more comprehensive view of a node's connectivity, capturing not only its direct neighbors but also its egonet structure. In addition, we also consider the size of a node's corresponding cluster, from herein after referred to as \textit{node size}. If a cluster is small in size, it must be either because its points belong to a relatively isolated interval of the open cover, which is a consequence of them being mapped to a region with low density by the filter function, or simply because the clustering algorithm does not detect similarities between these points and other points in the same interval pre-image. In either case, nodes with low node size contain data points that display a different behaviour from the rest, a key aspect to inform the anomaly detection mechanism. 

Finally, the node-level features calculated by the NetSimile feature extraction and the node sizes are aggregated at the customer level using a statistic (recall that a customer can belong to more than one node). The customers that belong to the bottom $n$-th percentile across all aggregated features belong to the most isolated nodes and are labeled as anomalous.
This methodology allows us to influence the proportion of anomalous customers that we wish to detect, a feature that can prove to be useful in many applications such as ours, through the choice of $n$. 

Let $M_\theta=(V,E)$ be a Mapper graph calculated using a parameter $\theta\in\Theta_S$. Suppose we have an algorithm \verb|find_customer_nodes|$(c)$ that, given a customer ID $c$ belonging to a list of customers IDs $C$, identifies the nodes of the Mapper graph to which the customer belongs. The pseudo-code for the anomaly detection algorithm is provided in Algorithm~\ref{alg:anomaly_detection}.

\algrenewcommand\algorithmicrequire{\textbf{Input:}}
\algrenewcommand\algorithmicensure{\textbf{Output:}}
\begin{algorithm}
\caption{Anomaly Detection on the Mapper Graph}
\label{alg:anomaly_detection}
\begin{algorithmic}
\Require Mapper graph \( M_\theta = (V, E) \), \( n \in \mathbb{N}_{\leq 100} \), customer IDs list \( C = \{c_1, \dots, c_m\} \)
\Ensure Subset \( C_{\text{anom}} \subseteq C \) of customers in the bottom \( n \)-th percentile of node feature means
\vspace{1mm}
\State Compute node-feature-matrix \( F \in \mathbb{R}^{|V| \times d} \gets \texttt{NetSimileFeatureExtraction}(M_\theta) \)
\State Append node sizes as an additional feature: \( F \gets \texttt{concat}(F, \mathbf{s}) \) where \( \mathbf{s} \in \mathbb{N}^{|V| \times 1} \)
\State Initialize customer-feature-matrix: \( X \in \mathbb{R}^{|C| \times (d+1)} \)
\For{each \( i, c \in \text{enumerate}(C) \)}
    \State \( V_c \gets \texttt{find\_customer\_nodes}(c) \subseteq V \)
    \State \( F_c \gets F[V_c, :] \in \mathbb{R}^{|V_c| \times (d+1)} \) \Comment{Feature vector of customer $c$}
    \State \( \mu_c \gets \frac{1}{|V_c|} \sum_{v \in V_c} F[v, :] \in \mathbb{R}^{d+1} \) \Comment{Mean feature vector}
    \State \( X[i, :] \gets \mu_c \)
\EndFor
\State Let \( \tau \in \mathbb{R}^{d+1} \) be the \( n \)-th percentile threshold for all columns of \( X \)
\State \( C_{\text{anom}} \gets \{c_i \in C \mid X[i,j] \leq \tau[j] \text{ for all } 1\leq j\leq d+1\} \)
\State \Return \( C_{\text{anom}} \)
\end{algorithmic}
\end{algorithm}

Beyond simply labeling customers as anomalous, this algorithm naturally provides an anomaly scoring method. Specifically, a customer $c$'s anomaly score can be derived by applying a statistical measure to summarize its mean vector $\mu_c$. This score can be used to rank anomalous customers and to apply further filtering.

\subsubsection{Ensemble Mapper Anomaly Detection}
\label{sec:ensemble_anomaly_detection}

The stability optimization routine (Section~\ref{sec:stability_optimization}) identifies a region of the parameter space that produces stable Mapper graphs. To reduce the variance in the anomaly detection algorithm's predictions, an ensemble learning technique was employed. Ensemble learning combines multiple models into a single predictive framework, enhancing robustness, stability, and overall performance~\cite{fitzpatrick2023ensemblelearningmapper}.  

In~\cite{kanglim2021ensemblemapper}, an ensemble learning for the Mapper graph has been proposed, showing promising results. However, integrating different Mapper graphs into a single structure is computationally complex and challenging to implement. In our anomaly detection task, this issue is avoided by performing ensemble learning without directly merging Mapper graphs, allowing us to bypass these complexities.  

Let $\theta_s \subseteq \Theta_S$ be a subspace of the stable region calculated through the stability optimization routine. For each $\theta \in \theta_s$, the Mapper graph $\mathcal{M}_\theta$ is constructed and the anomaly detection algorithm (Algorithm~\ref{alg:anomaly_detection}) is applied, producing binary labels $A_\theta$. The set of labels $\{A_\theta\}_{\theta \in \theta_s}$ is then aggregated using a majority vote: a customer is labeled as anomalous if and only if the majority of predictions $A_\theta$, $\theta \in \theta_s$, classify them as such~\cite{Rokach2010ensemblelearning}.

%%%%%%%%%%%%%%%%%%%%%%%%%%%%%%%%%%%%%%%%%%%
\subsection{Customer Segmentation: a brief overview}
\label{sec:customer_segmentation}

\label{sec:community_detection}
The network calculated by the Mapper algorithm is useful for more than just the anomaly detection task. Indeed, the graph's connectivity encodes information about inter-cluster relationships that can be exploited to understand complex community structures within and between these clusters. In the context of this paper, the identification and understanding of communities within the Mapper graph can be transformed into a segmentation of customers. In this section, we outline a procedure that exploits persistent homology to perform community detection on a connected mapper graph.

The goal of a community detection algorithm is to decipher complex structures and dynamics within graphs by clustering its nodes~\cite{li2024comprehensivereviewcommunitydetection}.  In our community detection algorithm, a mapper graph $M_\theta=(V,E)$ is endowed with a density function $f:V\rightarrow\mathbb{R}_+$, and the communities are identified by the most important density peaks. 
%This algorithm is motivated by the results which will be presented in Section~\ref{sec:results}, where we compare the communities detected by our approach with those detected by ToMATo~\cite{chazal2011}.

The steps of the customer segmentation procedure are the following.
\\\textbf{Step 1: Mapper Graph Construction.} Clusters of customers are represented as nodes, with edges informed by the AuToMATo clustering algorithm. The mapper graph is constructed using the most stable set of parameters as found by the stability optimization routine, and is endowed with a density function $f:V\rightarrow\mathbb{R}$.
\\\textbf{Step 2: Mode Identification.} Using persistent homology, the local density maxima are detected, representing cluster modes.
\\\textbf{Step 3: Weak Community Handling.} Similarly to ToMATo, our approach allows weak clusters (low-persistence modes) to merge into stronger modes via a cluster hierarchy, ensuring merges respect graph connectivity and topology. This yields a partition of the mapper graph nodes into communities.
\\\textbf{Step 4: Community Assignment.} Customers are assigned to the community their respective nodes belong to. A customer whose nodes belong to different communities will be assigned to the most frequent one.
%\end{itemize}

%Applied to the real-world financial dataset we study in this paper, the method effectively identifies communities of customers that display statistically different behaviour, as will be discussed in Section~\ref{sec:results_cust_seg}. 
While this procedure yields promising results (see  Section~\ref{sec:results_cust_seg}), ongoing work includes rigorous benchmarking against standard density-based methods.

\section{Results}
\label{sec:results}

In this section, we present the experimental results found by the approach presented in Section~\ref{sec:methodology}. We evaluated our methodology on a private banking dataset that contains transactions and customer information for approximately 1.4 million customers over one year. 

\subsection{Stability Optimization}

To conduct the stability-driven Mapper parameter selection process, a grid search across the following parameters was conducted: the open cover resolution $r$, the gain $g$, and the AuToMATo $k$ parameter. 

\begin{figure}
    \centering
    \includegraphics[width=\linewidth]{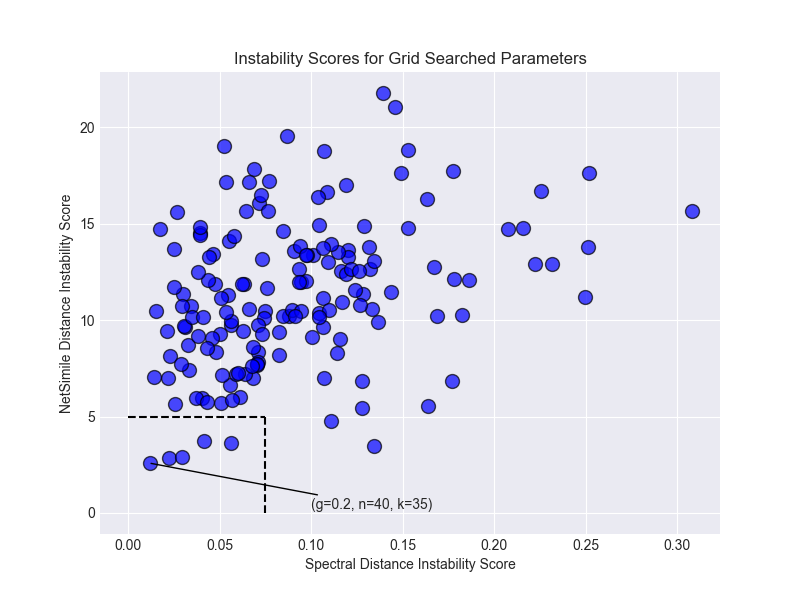}
    \caption{Results from the stability optimization grid search. $\theta_{opt}=(0.2, 40, 35)$ is shown on the plot as the most stable parameter value.}
    \label{fig:stability_optimization_grid}
\end{figure}

Figure~\ref{fig:stability_optimization_grid} presents the results of the stability optimization grid search, displaying the instability scores for the NetSimile distance and the spectral distance across various cover and clustering parameters. Each point in the plot represents a set of Mapper parameters $\theta=(g,n,k)$. The stability region $\Theta_S$ is highlighted in the bottom left quadrant.

%The parameter configuration with the lowest instability score, specifically $\theta_{opt} = (g, n, k) = (0.2, 40, 35)$, was selected as the optimal choice for performing customer segmentation and visualizing the anomaly detection algorithm. While this procedure prioritizes the stability of the Mapper graph, validating whether the outcome is robust for anomaly detection and customer segmentation requires further analysis.

\subsection{Anomaly Detection}

As explained in Section~\ref{sec:ensemble_anomaly_detection}, we select a subspace of the stable parameter region $\theta_s\subseteq \Theta_S$ to perform ensemble anomaly detection. This subspace was defined using the three parameter values that have the lowest instability score for both distance metrics, visible in Figure~\ref{fig:stability_optimization_grid}. 

\begin{figure}
    \centering
    \includegraphics[width=\linewidth]{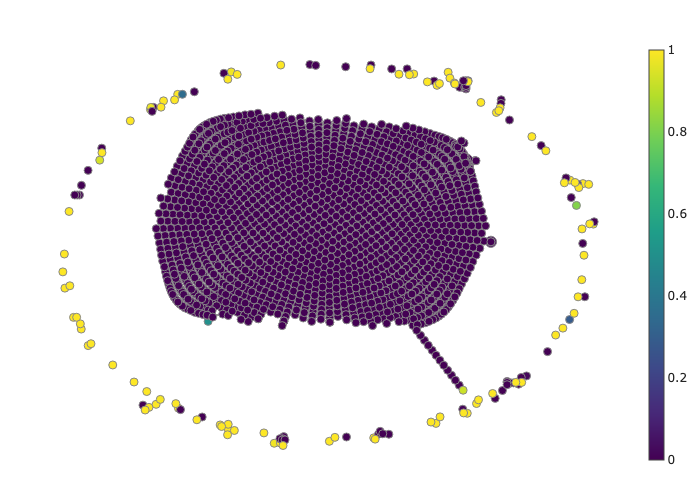}
    \caption{The Mapper graph with $\theta=\theta_{opt}$. The colour of the node indicates the proportion of anomalous customers present in the node.}
    \label{fig:anomalous_mapper}
\end{figure}

Figure~\ref{fig:anomalous_mapper} displays the Mapper graph calculated with parameters $\theta = \theta_{opt}=(0.2, 40, 35)$. The node colour varies depending on the proportion of anomalous customers detected in the node, with yellow indicating that all customers in the node are anomalous, and dark blue indicating that none are anomalous. The anomaly status of customers was determined using a majority vote on the output of Algorithm~\ref{alg:anomaly_detection} with three distinct Mapper graphs chosen from the stability region, and $n=10$.

The next two examples demonstrate the algorithm’s ability to detect diverse forms of anomalous financial behavior, a capability that is essential for effective unsupervised anomaly detection in real-world banking data.

\begin{table}[h]
    \centering    
    \caption{Transactional history of an anomalous customer}
    \begin{adjustbox}{width=\columnwidth}
    \begin{tabular}{ccccc}
        \toprule
        Date & Type & Amount Sent & Amount Received & Contraent Country \\
        \midrule
        05.07.2012 & transfer & 0.00 & 71.89 & Belgium \\
        11.07.2012 & transfer & 0.00 & 1110.00 & Austria \\
        11.07.2012 & transfer & 0.00 & 1674.07 & Austria \\
        13.07.2012 & transfer & 0.00 & 1672.52 & Austria \\
        16.07.2012 & transfer & 0.00 & 810.00 & Austria \\
        16.07.2012 & transfer & 0.00 & 1920.00 & Austria \\
        17.07.2012 & transfer & 0.00 & 54.00 & Estonia \\
        17.07.2012 & transfer & 0.00 & 64.99 & Austria \\
        17.07.2012 & transfer & 0.00 & 54.99 & Austria \\
        18.07.2012 & transfer & 0.00 & 1868.05 & Austria \\
        19.07.2012 & transfer & 0.00 & 1569.79 & Italy \\
        19.07.2012 & transfer & 0.00 & 98.36 & Austria \\
        20.07.2012 & transfer & 0.00 & 1968.05 & Austria \\
        24.07.2012 & transfer & 0.00 & 75.63 & Austria \\
        24.07.2012 & transfer & 0.00 & 104.49 & Austria \\
        25.07.2012 & transfer & 0.00 & 49.99 & Austria \\
        30.07.2012 & transfer & -20999.51 & 0.00 & China \\
        30.07.2012 & transfer & 0.00 & 124.99 & Italy \\
        \bottomrule
    \end{tabular}
    \end{adjustbox}
    \label{tab:anomaly_example_1}
\end{table}

The customer whose transactions are contained in Table \ref{tab:anomaly_example_1} received a large number of high-value inbound transfers over a short time span, primarily from Austria, without sending any funds until a single, exceptionally large outbound transfer to China at the end of the month. This sharp asymmetry between inbound and outbound flows, combined with the abrupt spike in outflow to a non-European country, deviates significantly from typical customer behaviour observed in the dataset. This behaviour is consistent with known red flags in financial crime, e.g. \emph{layering} or \emph{money mule} activity. 

\begin{table}[h]
    \centering
    \caption{Transactional history of an anomalous customer}
    \begin{adjustbox}{width=\columnwidth}
    \begin{tabular}{ccccc}
        \toprule
        Date & Type & Amount Sent & Amount Received & Contraent Country \\
        \midrule
        02.07.2012 & commission & -18.20 & 0.00 & Germany \\
        02.07.2012 & ATM & -30.00 & 0.00 & Germany \\
        18.07.2012 & transfer & 0.00 & 104.96 & Germany \\
        18.07.2012 & transfer & 0.00 & 110.08 & Germany \\
        18.07.2012 & transfer & 0.00 & 115.20 & Germany \\
        18.07.2012 & transfer & 0.00 & 189.44 & Germany \\
        18.07.2012 & transfer & 0.00 & 10.24 & Germany \\
        18.07.2012 & transfer & 0.00 & 115.20 & Germany \\
        18.07.2012 & transfer & 0.00 & 151.04 & Germany \\
        18.07.2012 & transfer & 0.00 & 28.16 & Germany \\
        18.07.2012 & transfer & 0.00 & 15.36 & Germany \\
        18.07.2012 & transfer & 0.00 & 332.66 & Germany \\
        18.07.2012 & transfer & 0.00 & 250.61 & Germany \\
        18.07.2012 & transfer & 0.00 & 107.52 & Germany \\
        18.07.2012 & transfer & 0.00 & 40.96 & Germany \\
        18.07.2012 & transfer & 0.00 & 112.64 & Germany \\
        18.07.2012 & transfer & 0.00 & 17.92 & Germany \\
        18.07.2012 & transfer & 0.00 & 278.88 & Germany \\
        18.07.2012 & transfer & 0.00 & 163.84 & Germany \\
        18.07.2012 & transfer & 0.00 & 208.34 & Germany \\
        23.07.2012 & ATM & -100.00 & 0.00 & Germany \\
        25.07.2012 & ATM & -500.00 & 0.00 & Germany \\
        27.07.2012 & ATM & -100.00 & 0.00 & Germany \\
        20.08.2012 & transfer & 0.00 & 264.44 & Germany \\
        \bottomrule
    \end{tabular}
    \end{adjustbox}
    \label{tab:anomaly_example_2}
\end{table}

Another example highlighting the strengths of the algorithm is a customer whose behaviour deviates in a different but equally suspicious manner. This customer, shown in Table~\ref{tab:anomaly_example_2}, received over twenty small inbound transfers on a single day, and then engaged in multiple cash withdrawals shortly afterward. The large volume and density of inbound transfers, especially when combined with immediate withdrawals and the absence of substantial outbound transfers to other accounts, may be indicative of \emph{smurfing}, a  technique used to evade financial reporting thresholds.

\subsection{Customer Segmentation}
\label{sec:results_cust_seg}

The input Mapper graph used for the customer segmentation procedure was constructed using the most stable parameter $\theta=\theta_{opt}=(0.2,40,35)$. After applying the anomaly detection algorithm and removing the detected anomalies, we endowed the mapper graph nodes with a density function $f$ that maps graph nodes to their respective sizes.

\begin{figure}
    \centering
    \includegraphics[width=\linewidth]{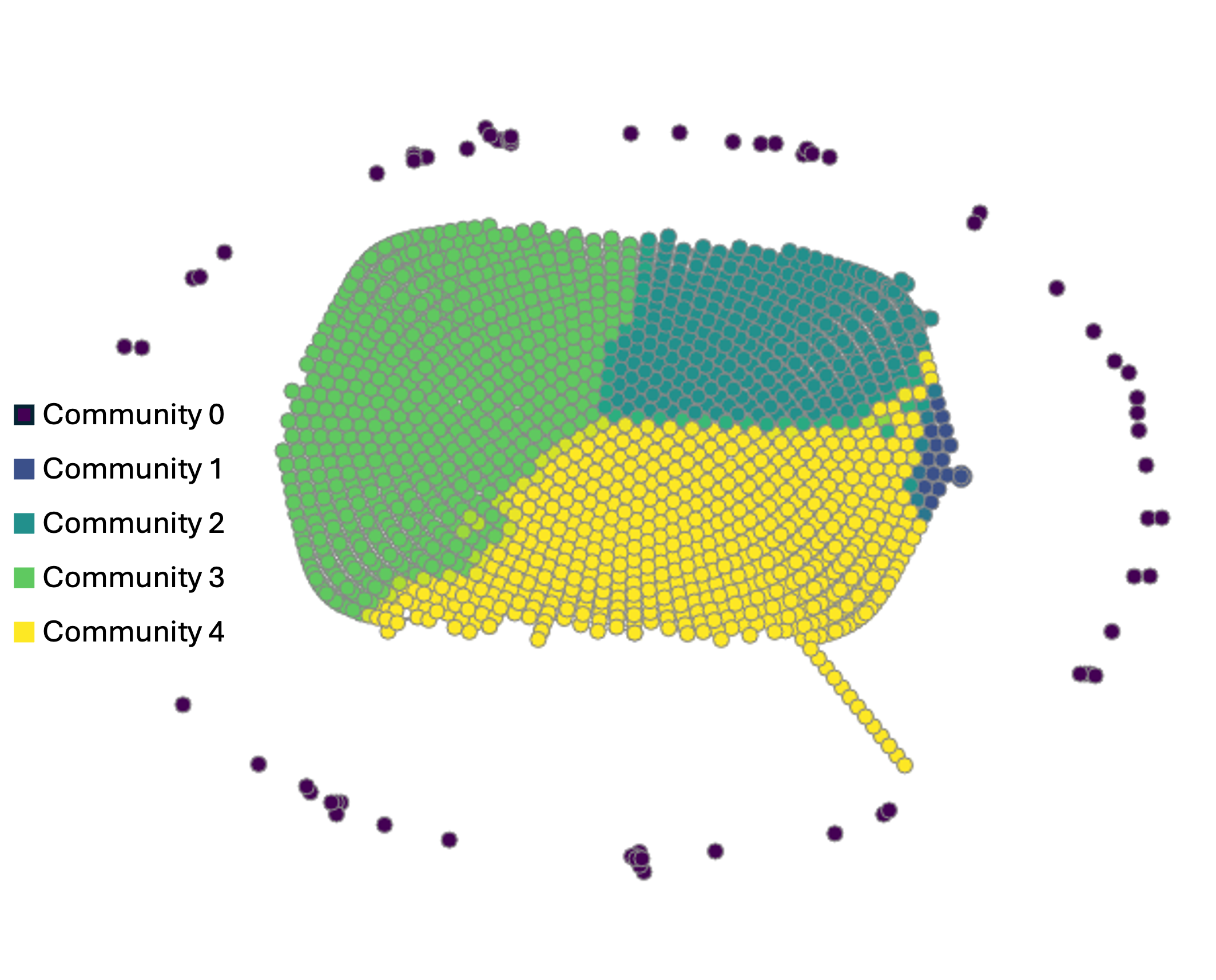}
    \caption{Communities of the Mapper graph detected by our Algorithm. Each colour represents a different community.}
    \label{fig:communities_plot}
\end{figure}

Figure~\ref{fig:communities_plot} shows the communities detected on the Mapper graph using our algorithm. It is important to highlight that our community detection algorithm is applied to a connected graph. In this case, we focused solely on the connected component located at the center of the plot. The Mapper nodes scattered around the plot's border represent nodes that were not classified as anomalous by the anomaly detection algorithm and were excluded from the community detection process. Consequently, these nodes were treated as a separate community, labeled Community 0. As a result, a total of 5 communities were identified, labeled from 0 to 4 (see the colour legend of Figure~\ref{fig:communities_plot}).

The adeptness of the customer segmentation yielded by our approach was assessed by measuring whether the separation between any two communities is statistically significant. This was done using Permutational Multivariate Analysis of Variance (PERMANOVA)~\cite{anderson2001permanova}. To control for false discoveries, a p-value correction method, the False Discovery Rate (FDR), was used to ensure that the results are statistically robust and not due to random chance~\cite{benjamini1995fdr}. 

\begin{table}[h]
    \centering
    \caption{Corrected p-values and significance results for pairwise comparisons of customer communities.}
    \begin{adjustbox}{width=\columnwidth}
    \begin{tabular}{cccc}
        \toprule
        Community A & Community B & Corrected p-value & Significance \\
        \midrule
        0 & 1 & 0.0014 & True \\
        0 & 2 & 0.1970 & False \\
        0 & 3 & 0.0014 & True \\
        0 & 4 & 0.0014 & True \\
        1 & 2 & 0.1522 & False \\
        1 & 3 & 0.0014 & True \\
        1 & 4 & 0.0014 & True \\
        2 & 3 & 0.0014 & True \\
        2 & 4 & 0.0014 & True \\
        3 & 4 & 0.0075 & True \\
        \bottomrule
    \end{tabular}
    \end{adjustbox}
    \label{tab:significance_results}
\end{table}

Table~\ref{tab:significance_results} displays the results of the FDR-corrected pairwise PERMANOVA tests between each of the communities shown in Figure~\ref{fig:communities_plot}. The $p$-value to determine significance was set at $p=0.05$.

Most communities are found to be significantly different from the others. The only exceptions occur in the comparisons between communities 2 and 0, and between communities 2 and 1. The first exception can be attributed to the nature of the definition of community 0, as this community is not detected by the community detection algorithm but rather grouped together due to its lack of connectivity to the rest of the graph. Regarding the insignificance of the comparison between communities 1 and 2, further investigation is needed, such as incorporating additional behavioural features or refining the community detection approach.

Compared to ToMATo, our community detection algorithm was found to be better suited to perform customer segmentation. The same statistical analysis presented above was used to assess the customer segments detected by ToMATo on the Mapper graph, yielding the poor results shown in Table~\ref{tab:tomato_significance_results}. 
%Indeed, most community comparisons demonstrated statistical insignificance, unlike our approach.

\begin{table}[h]
    \centering
    \caption{Corrected p-values and significance results for pairwise comparisons of customer communities using ToMATo.}
    \begin{adjustbox}{width=\columnwidth}
    \begin{tabular}{cccc}
        \toprule
        Community A & Community B & Corrected p-value & Significance \\
        \midrule
        0 & 1 & 0.2677 & False \\
        0 & 2 & 0.0030 & True \\
        0 & 3 & 0.0030 & True \\
        0 & 4 & 0.2677 & False \\
        1 & 2 & 0.7896 & False \\
        1 & 3 & 0.8860 & False \\
        1 & 4 & 0.2677 & False \\
        2 & 3 & 0.0857 & False \\
        2 & 4 & 0.0030 & True \\
        3 & 4 & 0.0030 & True \\
        \bottomrule
    \end{tabular}
    \label{tab:tomato_significance_results}
    \end{adjustbox}
\end{table}

\section{Conclusions and Future Work}
This paper demonstrated the utility of Topological Data Analysis for uncovering structure in high‑dimensional, unlabeled banking data. We developed a stability‑driven parameter selection method and an unsupervised anomaly detector based on graph connectivity. These tools succeded in revealing money mule and smurfing cases, highlighting the Mapper graph’s expressiveness in capturing diverse financial irregularities. We also briefly outlined a persistence‑driven community detection approach that is highly compatible with the anomaly detector, and shows encouraging results. 

Despite these promising outcomes, key limitations point to avenues for further research. First, the anomaly detector’s sensitivity to the choice of filter function suggests a systematic study of alternative functions to understand their influence on detection performance. Second, the community detection method, while often yielding statistically significant segment separations, requires rigorous comparison against other graph and topology based clustering techniques. 

Potential applications include exploring the interpretability of topological features in the mapper graph, linking nodes or cycles to concrete fraud typologies, which may yield actionable insights for compliance and risk teams. Extending these TDA‑based methods to real‑time detection or other domains would test their scalability and robustness, paving the way for broader applications of unsupervised, high‑dimensional anomaly detection.

\section*{Acknowledgments}
This project was part of a Master thesis at ETH Zürich, under the supervision of Dr.\ Patrick Schnider and Dr.\ Linda Maria De Cave.

\section*{CRediT authorship contribution statement}
\textbf{Leonardo Aldo Alejandro Barberi:} conceptualization, data curation, formal analysis, investigation, methodology, validation, visualization, writing - original draft.
\\\textbf{Linda Maria De Cave:} project administration, resources, supervision, writing - review and editing.

%% The Appendices part is started with the command \appendix;
%% appendix sections are then done as normal sections

%% If you have bib database file and want bibtex to generate the
%% bibitems, please use
%%
%%  \bibliographystyle{elsarticle-num} 
%%  \bibliography{<your bibdatabase>}

\begin{thebibliography}{99}

%% For numbered reference style
%% \bibitem{label}
%% Text of bibliographic item

%\bibitem{bolton2002}
%R. J. Bolton and D. J. Hand, ``Statistical fraud detection: A review,'' \emph{Statistical Science}, vol. 17, no. 3, pp. 235–255, 2002. 

\bibitem{bahnsen2016}
A. C. Bahnsen, D. Aouada, A. Stojanovic, and B. Ottersten, Feature engineering strategies for credit card fraud detection, Expert Systems with Applications, vol. 51, pp. 134–142, 2016. https://doi.org/10.1016/j.eswa.2015.12.030

\bibitem{dalpozzo2015}
A. Dal Pozzolo, O. Caelen, Y. Le Borgne, S. Waterschoot, and G. Bontempi, Credit card fraud detection: A realistic modeling and a novel learning strategy, IEEE Transactions on Neural Networks and Learning Systems, vol. 29, no. 8, pp. 3784–3797, 2018. https://doi.org.10.1109/TNNLS.2017.2736643

%\bibitem{kou2004}
%Y. Kou, C.-T. Lu, S. Sirwongwattana, and Y.-P. Huang, ``Survey of fraud detection techniques,'' in \emph{Proceedings of the IEEE International Conference on Networking, Sensing and Control}, 2004, pp. 749–754. 

%\bibitem{phua2010}
%C. Phua, V. Lee, K. Smith, and R. Gayler, ``A comprehensive survey of data mining-based fraud detection research,'' \emph{arXiv preprint}, arXiv:1009.6119, 2010. 

\bibitem{jurgovsky2018}
J. Jurgovsky, M. Granitzer, S. Ziegler, L. Calabretto, and D. Portier, Sequence classification for credit-card fraud detection, Expert Systems with Applications, vol. 100, pp. 234–245, 2018. https://doi.org/10.1016/j.eswa.2018.01.037

\bibitem{ngai2011}
E. W. T. Ngai, Y. Hu, Y. H. Wong, Y. Chen, and X. Sun, The application of data mining techniques in financial fraud detection: A classification framework and an academic review of literature, Decision Support Systems, vol. 50, no. 3, pp. 559–569, 2011. https://doi.org/10.1016/j.dss.2010.08.006

\bibitem{bhattacharyya2011}
S. Bhattacharyya, S. Jha, K. Tharakunnel, and J. C. Westland, Data mining for credit card fraud: A comparative study, Decision Support Systems, vol. 50, no. 3, pp. 602–613, 2011. https://doi.org/10.1016/j.dss.2010.08.008


\bibitem{mozumder2024optimizing}
Md A. S. Mozumder, F. Mahmud, M. S. Shak, and M. S. M. Bhuiyan, Optimizing Customer Segmentation in the Banking Sector: A Comparative Analysis of Machine Learning Algorithms, Journal of Computer Science and Technology Studies, vol. 6, no. 4, pp. 01-07, 2024. https://al-kindipublisher.com/index.php/jcsts/article/view/7836

\bibitem{John2023}
John, J.M., Shobayo, O., and Ogunleye, B. (2023), An Exploration of Clustering Algorithms for Customer Segmentation in the UK Retail Market, Analytics, vol. 2, pp. 809–823, 2023. https://doi.org/10.3390/analytics2040042

%\bibitem{Nasirzonouzi2025}
%Nasirzonouzi, A., Leveraging Network Science for Customer Segmentation and Product Recommendation, Northeast Journal of Complex Systems, vol. 7, article 3, 2025. https://doi.org/10.22191/nejcs/vol7/iss1/3

%\bibitem{riveracastro2020topologyclusterregression}
%R. Rivera‑Castro, A. Pletnev, P. Pilyugina, G. Diaz, I. Nazarov, W. Zhu, and E. Burnaev,  
%Topology-based Clusterwise Regression for User Segmentation and Demand Forecasting,  
%Expert Systems with Applications, vol. 160, 113641, 2020.  
%https://doi.org/10.1016/j.eswa.2020.113641

\bibitem{singh2007topological}
G. Singh, F. Mémoli, and G. Carlsson,
Topological Methods for the Analysis of High Dimensional Data Sets and 3D Object Recognition,
Eurographics Symposium on Point-Based Graphics, 2007, pp. 91--100. https://doi.org/10.2312/SPBG/SPBG07/091-100

\bibitem{edelsbrunner2010computational}
H. Edelsbrunner and J. Harer,
Computational Topology: An Introduction,
American Mathematical Society, 2010. https://doi.org/10.1007/978-3-540-33259-6\_7

\bibitem{binshiraj2024topologicalanomalydetection}
M. M. B. Shiraj, M. M. Rahman, M. Al-Imran, M. Z. A. Liza, M. M. Murshed, and N. Akhter,  
Anomaly detection in financial time series data via mapper algorithm and DBSCAN clustering,  
World Journal of Advanced Engineering Technology and Sciences, vol. 13, no. 1, 2024. https://doi.org/10.30574/wjaets.2024.13.1.0396

\bibitem{Rokach2010ensemblelearning}
L. Rokach,  
Ensemble-based classifiers,  
Artificial Intelligence Review, 2010. https://doi.org/10.1007/s10462-009-9124-7

\bibitem{fitzpatrick2023ensemblelearningmapper}
P. Fitzpatrick, A. Jurek-Loughrey, P. Dlotko, and J. M. Del Rincon,  
Ensemble Learning for Mapper Parameter Optimization,  
2023 IEEE 35th International Conference on Tools with Artificial Intelligence (ICTAI), 2023, pp. 129--134. https://doi.org/10.1109/ICTAI59109.2023.00026

\bibitem{kanglim2021ensemblemapper}
S. J. Kang and Y. Lim,  
Ensemble mapper,  
Stat, vol. 10, no. 1, pp. e405, 2021. https://doi.org/10.1002/sta4.405

\bibitem{oulhaj2024differentiablemappertopologicaloptimization}
Z. Oulhaj, M. Carrière, and B. Michel,  
Differentiable Mapper For Topological Optimization Of Data Representation,  
arXiv preprint arXiv:2402.12854, 2024. https://arxiv.org/abs/2402.12854

\bibitem{carrière2017statisticalanalysisparameterselection}
M. Carrière, B. Michel, and S. Oudot,  
Statistical Analysis and Parameter Selection for Mapper,  
arXiv preprint arXiv:1706.00204, 2017. https://arxiv.org/abs/1706.00204

\bibitem{byrne-etal-2022-topic}
C. Byrne, D. Horak, K. Moilanen, and A. Mabona,  
Topic Modeling With Topological Data Analysis,  
Proceedings of the 2022 Conference on Empirical Methods in Natural Language Processing, 2022, pp. 11514--11533. https://doi.org/10.18653/v1/2022.emnlp-main.792

\bibitem{ballester2024expressivitypersistenthomologygraph}
R. Ballester and B. Rieck,  
On the Expressivity of Persistent Homology in Graph Learning,  
arXiv preprint arXiv:2302.09826, 2024. https://arxiv.org/abs/2302.09826

\bibitem{chazal2011}
Frédéric Chazal, Leonidas J. Guibas, Steve Y. Oudot, and Primoz Skraba, Persistence-based clustering in Riemannian manifolds, Journal of the ACM, vol. 60, no. 6, pp. 1-38, 2011. https://doi.org/10.1145/2535927

\bibitem{huber2024automatooutoftheboxpersistencebasedclustering}
M. Huber, S. Kalisnik, and P. Schnider,  
AuToMATo: An Out-Of-The-Box Persistence-Based Clustering Algorithm,  
arXiv preprint arXiv:2408.06958, 2024. https://arxiv.org/abs/2408.06958

\bibitem{berlingerio2012netsimilescalableapproachsizeindependent}
M. Berlingerio, D. Koutra, T. Eliassi-Rad, and C. Faloutsos,  
NetSimile: A Scalable Approach to Size-Independent Network Similarity,  
arXiv preprint arXiv:1209.2684, 2012. https://arxiv.org/abs/1209.2684

\bibitem{li2024comprehensivereviewcommunitydetection}
J. Li, S. Lai, Z. Shuai, Y. Tan, Y. Jia, M. Yu, Z. Song, X. Peng, Z. Xu, Y. Ni, H. Qiu, J. Yang, Y. Liu, and Y. Lu,  
A Comprehensive Review of Community Detection in Graphs,  
arXiv preprint arXiv:2309.11798, 2024. https://arxiv.org/abs/2309.11798


\bibitem{anderson2001permanova}
M. Anderson,  
A new method for non-parametric multivariate analysis of variance,  
Austral Ecology, vol. 26, pp. 32--46, 2001. https://doi.org/10.1111/j.1442-9993.2001.01070.pp.x

\bibitem{benjamini1995fdr}
Y. Benjamini and Y. Hochberg,  
Controlling the False Discovery Rate: A Practical and Powerful Approach to Multiple Testing,  
Journal of the Royal Statistical Society. Series B (Methodological), vol. 57, no. 1, pp. 289--300, 1995. http://www.jstor.org/stable/2346101


\end{thebibliography}

%% else use the following coding to input the bibitems directly in the
%% TeX file.

%% Refer following link for more details about bibliography and citations.
%% https://en.wikibooks.org/wiki/LaTeX/Bibliography_Management

\end{document}